\documentclass[10pt,twocolumn,letterpaper]{article}

\usepackage{cvpr}              
\definecolor{cvprblue}{rgb}{0.21,0.49,0.74}
\usepackage[pagebackref,breaklinks,colorlinks,allcolors=cvprblue]{hyperref}

\usepackage{amsthm,amsmath,amssymb}
\usepackage{mathrsfs}
\usepackage{algorithm}
\usepackage{algpseudocode}
\usepackage{multirow}

\title{ParaUni: Enhance Generation in Unified Multimodal Model with Reinforcement-driven Hierarchical Parallel Information Interaction}

\author{Jiangtong Tan\textsuperscript{1,2}, Lin Liu\textsuperscript{1,2}, Jie Huang\textsuperscript{1}\footnotemark[2], Xiaopeng Zhang\textsuperscript{2}, Qi Tian\textsuperscript{2}\footnotemark[1], Feng Zhao\textsuperscript{1}\footnotemark[1]\\
\textsuperscript{1}MoE Key Laboratory of Brain-inspired Intelligent Perception and Cognition,\\
University of Science and Technology of China\\
\textsuperscript{2}Huawei Inc.\\
{\tt\small \{jttan, ll0825, hj0117\}@mail.ustc.edu.cn, \{zhangxiaopeng12, tian.qi1\}@huawei.com} \\
{\tt\small
fzhao956@ustc.edu.cn}
}

\begin{document}
\let\oldtwocolumn\twocolumn
\renewcommand\twocolumn[1][]{
    \oldtwocolumn[{#1}{
    \begin{center}
    \includegraphics[width=0.8\textwidth]{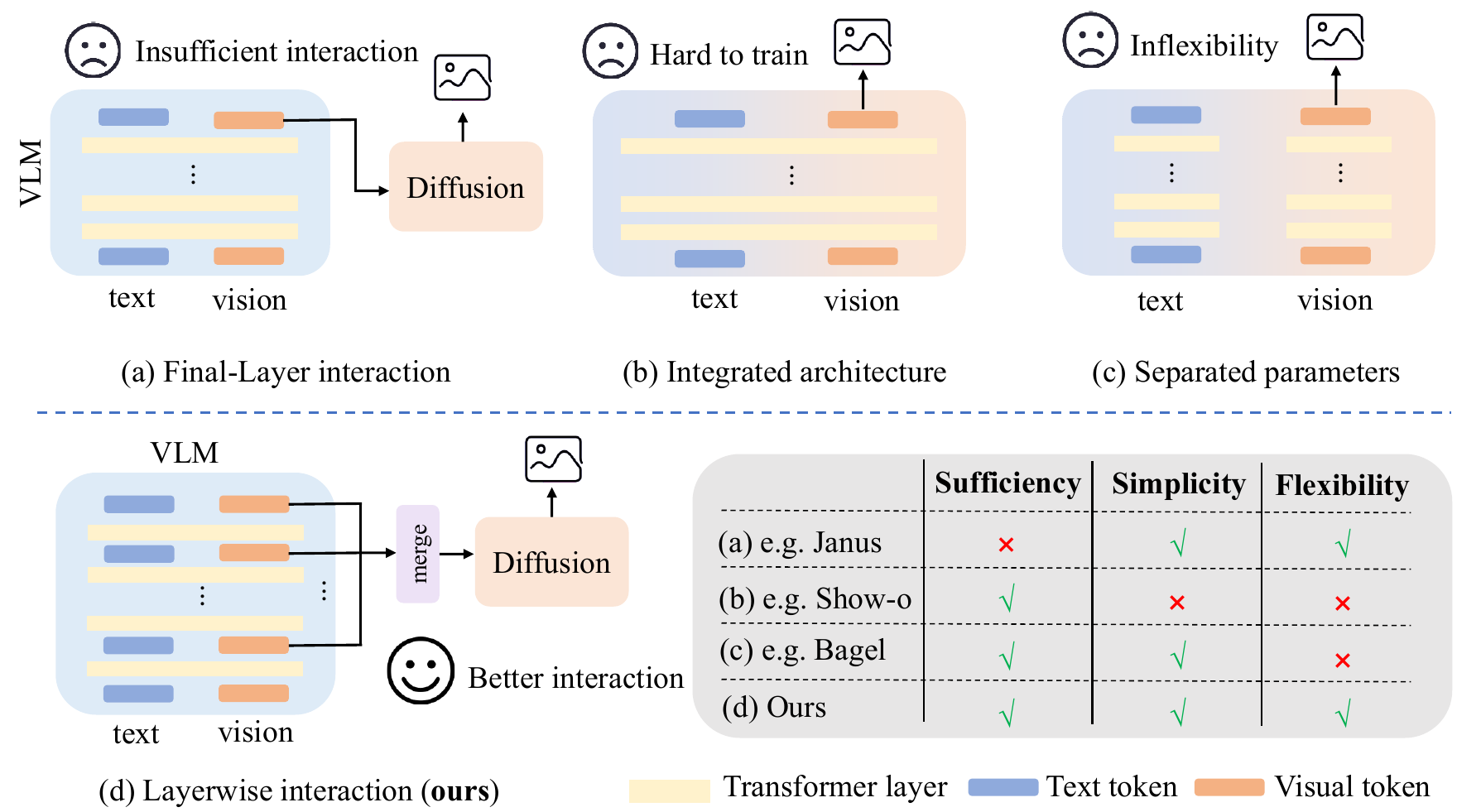}
    \captionof{figure}{Illustration of different VLM and diffusion interaction methods in unified understanding and generation. (a) Final-Layer interaction, which only uses the features from the last layer of the VLM as a condition, is insufficient in capturing detailed image contents and only provides a semantic abstraction. (b) Integrated architecture, which integrates VLM and diffusion into one transformer, faces increased training difficulty due to different training objectives. (c) Separated parameters, which use two sets of parameters for understanding and generating information within one transformer, although beneficial for multimodal information interaction, faces challenges in flexibility and scalability. (d) Our layerwise interaction, which merges different levels of semantic abstraction and fine-grained details from the VLM, ensures sufficiet interaction and retains  flexible separation architecture.}
    \label{fig:mainapp}
    \end{center}
    }]
}
\maketitle
\renewcommand{\thefootnote}{\fnsymbol{footnote}} 
\footnotetext[1]{Corresponding author}
\footnotetext[2]{Technique contributor}
\renewcommand{\thefootnote}{\arabic{footnote}}  
\begin{abstract}
\label{sec:abs}
Unified multimodal models significantly improve visual generation by combining vision-language models (VLMs) with diffusion models. However, existing methods struggle to fully balance sufficient interaction and flexible implementation due to vast representation difference. Considering abundant and hierarchical information in VLM's layers from low-level details to high-level semantics, we propose \textbf{ParaUni}. It extracts features from variants VLM's layers in a \textbf{Para}llel way for comprehensive information interaction and retains a flexible separation architecture to enhance generation in \textbf{Uni}fied multimodal model. Concretely, visual features from all VLM's layers are fed in parallel into a Layer Integration Module (LIM), which efficiently integrates fine-grained details and semantic abstractions and provides the fused representation as a condition to the diffusion model. To further enhance performance, we reveal that these hierarchical layers respond unequally to different rewards in Reinforcement Learning (RL). Crucially, we design a Layer-wise Dynamic Adjustment Mechanism (LDAM) to facilitate multiple reward improvements that aligns the hierarchical properties of these layers using RL. Extensive experiments show ParaUni leverages complementary multi-layer features to substantially improve generation quality and shows strong potential for multiple reward advances during RL stages. Code is available at \href{https://github.com/JosephTiTan/ParaUni}{https://github.com/JosephTiTan/ParaUni}.
\end{abstract}

\section{Introduction}
\label{sec:intro}

As unified multimodal understanding and generation has long been a research focus, extensive work has demonstrated strong performance of unified architectures ~\cite{team2024chameleon, wu2024vila, xie2024show, wu2024janus, li2024synergen, deng2025bagel}, improving tasks such as image generation and editing.  However, due to the architectural gap between autoregressive (AR)-based vision-language models (VLMs) for image understanding and diffusion-based models for image generation,  unifying them in a single framework is highly challenging. 

Several prior works~\cite{zhang2025nexus, liu2025step1x} attempt to bridge this gap by supplying the final layer representations of a VLM as conditioning to a diffusion decoder, as shown in \cref{fig:mainapp} (a). However, reliance on insufficient information interaction from only a single layer restricts the fidelity and granularity of information transferred from the understanding module to the generation module, thus constraining potential gains in generation quality, which has been shown in \cite{li2025unifusion}. Alternative approaches \cite{xie2024show, zhou2024transfusion, zhao2024monoformer} seek to integrate the diffusion denoising process with the AR process within a single transformer to facilitate interaction, as shown in \cref{fig:mainapp}(b). Since the two procedures are governed by disparate optimization objectives, this integration substantially increases training complexity and precludes straightforward reuse of off-the-shelf pretrained models. Some works~\cite{shi2024llamafusion} use two separate transformer experts to handle understanding and generation,  while tokens from different modalities interact via shared multimodal self-attention inside each transformer block, as shown in \cref{fig:mainapp} (c). This design enables richer cross‑modal information interaction and reduces the difficulty of multi-task learning~\cite{deng2025bagel}. However, it is limited in flexibility and scalability and incurs high inference latency due to the tight coupling of the two type parameters~\cite{chen2025blip3o}.

\begin{figure}[t]
  \centering
  \includegraphics[width=1\linewidth]{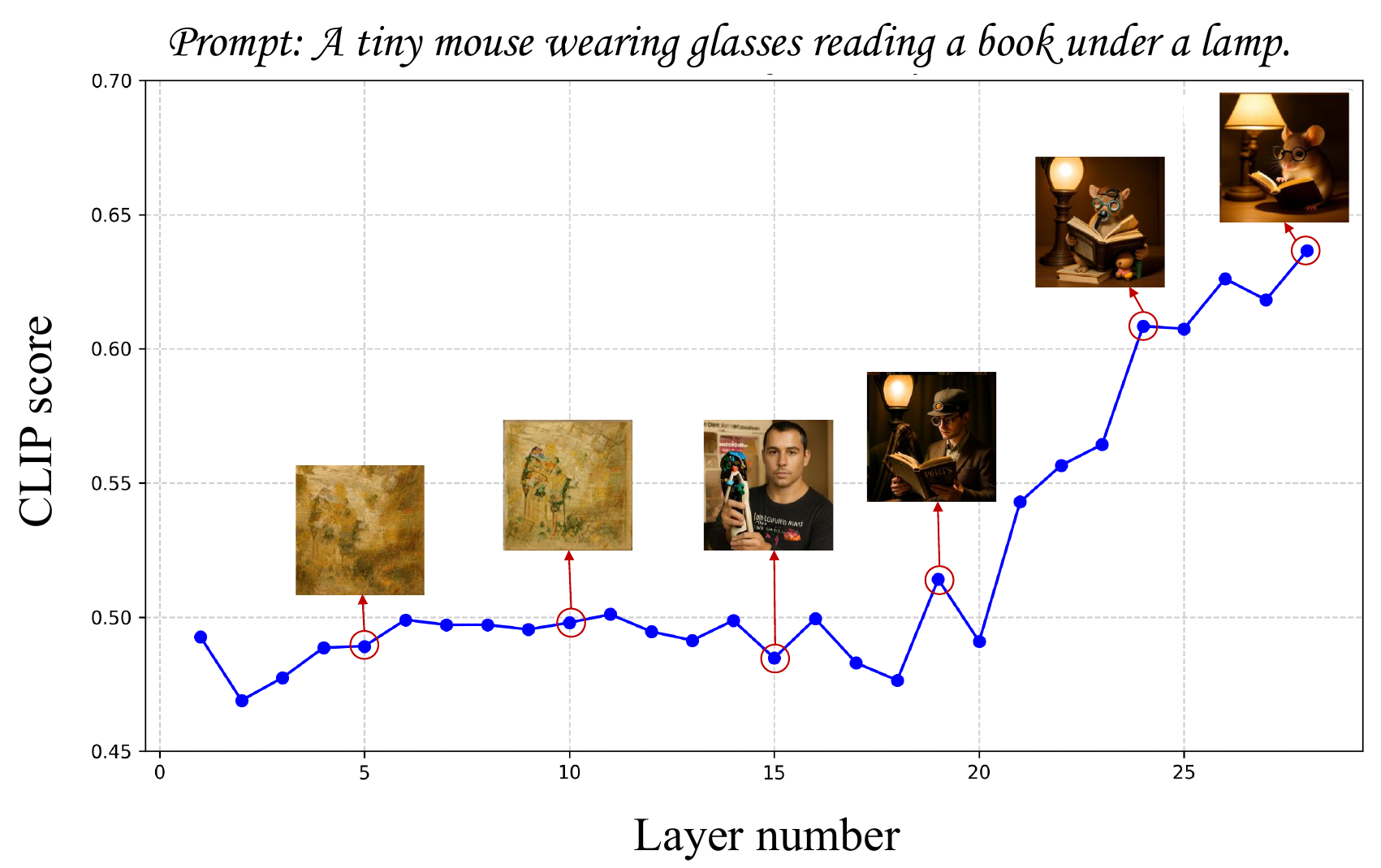}
   \caption{Illustration of CLIP score changes with 28 layer features from shallow to deep in VLM. We conduct experiments on 500 prompts and find that as the number of VLM layers increased, the images generated by the diffusion model gradually shifted from focusing on detailed textures to enhancing semantic information. The CLIP score, which measures the alignment between images and text, also increased with the number of layers. It indicates that different depths of the VLM's transformer layers encode information ranging from low-level details to high-level semantics.}
   \label{fig:top}
\end{figure}

\begin{figure}[t]
  \centering
  \includegraphics[width=1\linewidth]{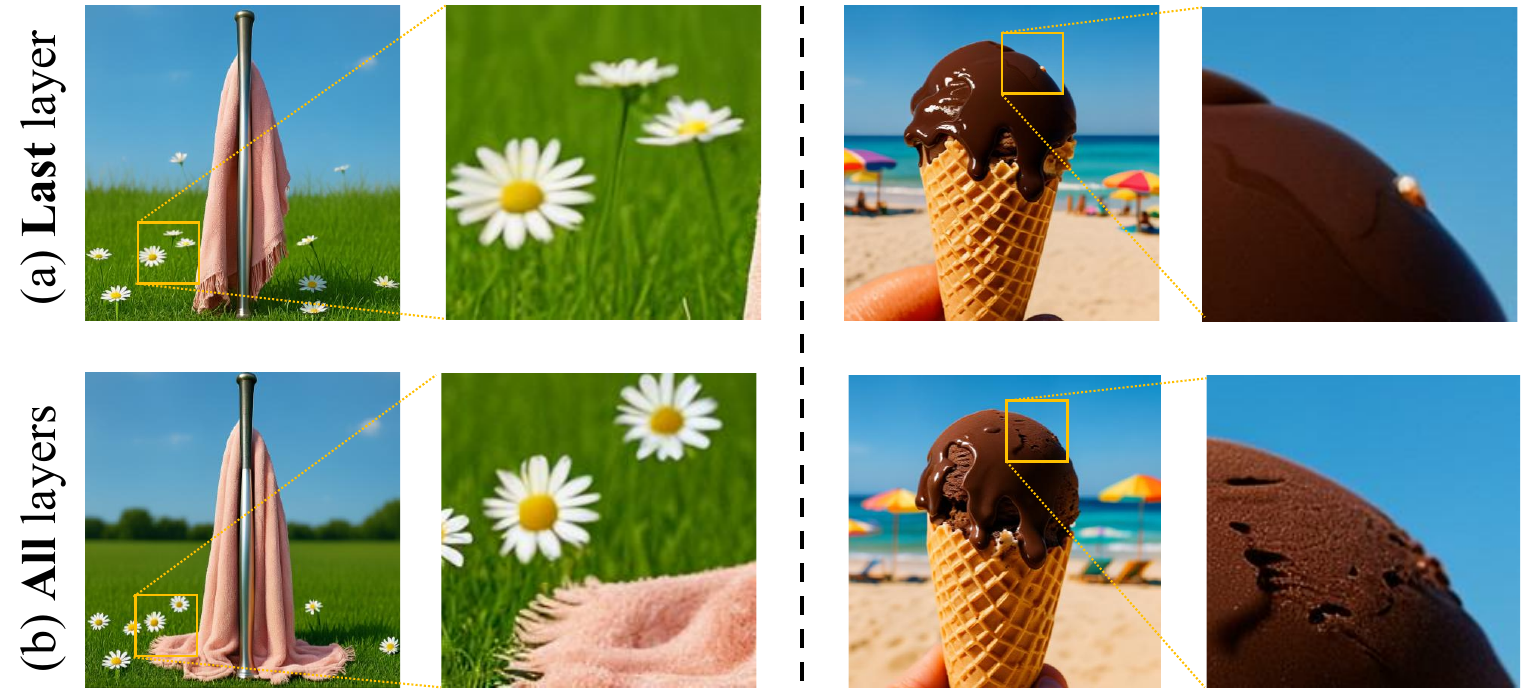}
   \caption{Comparison between (a) using only the last layer of the VLM and (b) using all layers. After using all layers, the generated images have more details, indicating that the diffusion model integrates more detailed information from the VLM, which confirms the rationality of using all layers as conditions.}
   \label{fig:com}
\end{figure}

Prior works \cite{durrani2020analyzing, hennigen2020intrinsic, durrani2023discovering, lam2024large} have established that VLM's transformer layers at different depths encode varying levels of information from low-level details to high-level semantics. We argue that aggregating these features as conditions for generation with a diffusion model can effectively improve comprehensive interaction between them. To prove it, we extract the visual tokens separately from \textbf{each} layer of the VLM and use them as conditioning inputs to the diffusion model. We evaluate CLIP scores of generated images to reflect their semantic alignment \cite{radford2021learning, lee2022uniclip}. As shown in \cref{fig:top}, generated images conditioned on shallow layers tend to exhibit texture-like characteristics. As layer depth increases, the generated images progressively exhibit stronger semantic content, and their content becomes more consistent with the text prompt. From this analysis we conclude that VLM layers exhibit different properties from shallow to deep, so integrating all layers naturally provides more comprehensive information for the generation process. Therefore, we aggregate the visual tokens from \textbf{all} layers and feed the combined condition into the diffusion model during training. As illustrated in \cref{fig:com}, incorporating multi-layer information into the conditioning yields images with richer, more realistic details and further improves performance.

\begin{figure}[t]
  \centering
  \includegraphics[width=1\linewidth]{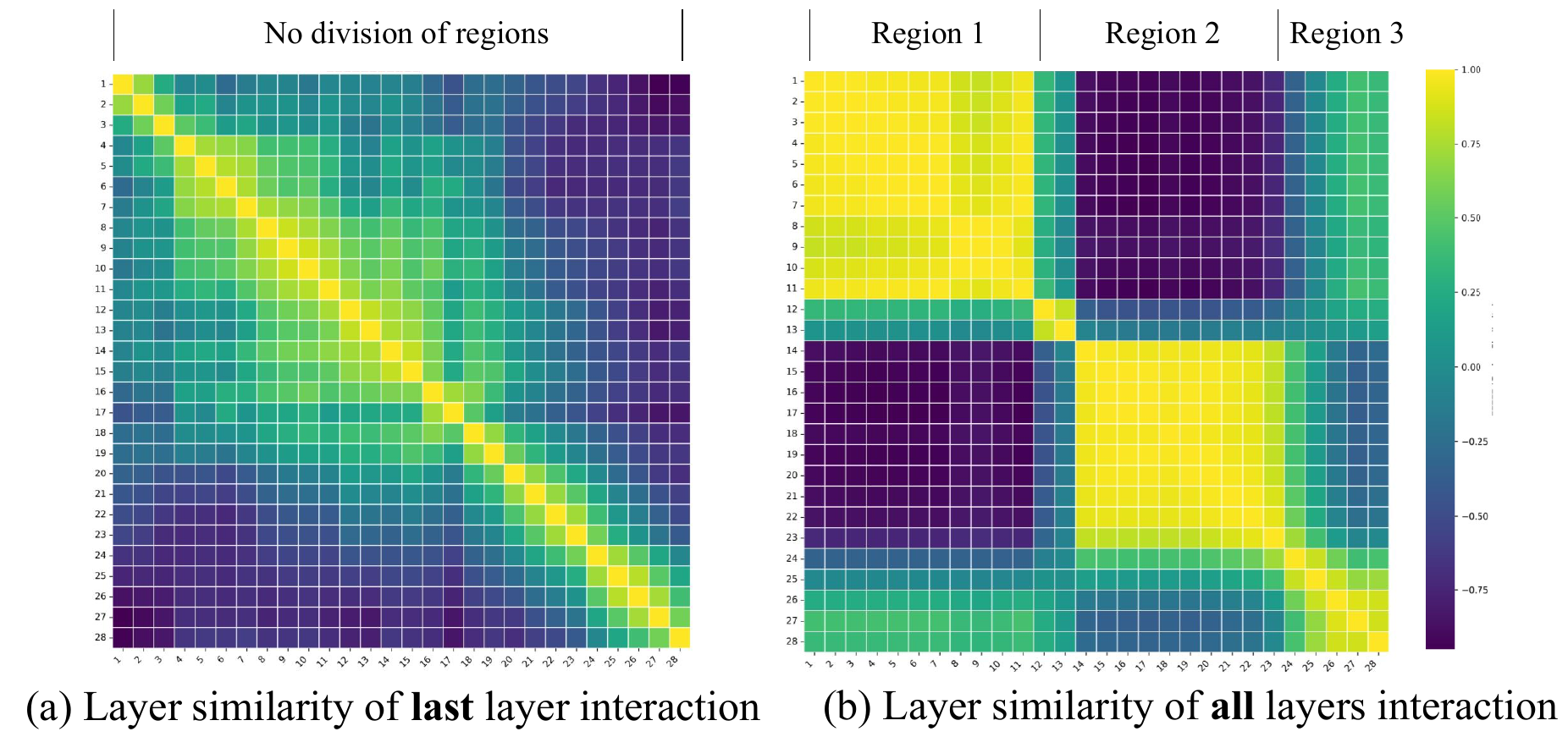}
   \caption{Layer similarity of the last layer or all layer interaction. For last-layer interaction, the similarity between layers is relatively low, while for all-layer interaction, cluster-like phenomena occur between layers. From \cite{li2025unifusion}, the phenomenon is not limited to our base model with universality. We analyze the properties of each clustered region in \cref{fig:reward}.}
   \label{fig:sim}
\end{figure}

\begin{figure}[t]
  \centering
  \includegraphics[width=0.85\linewidth]{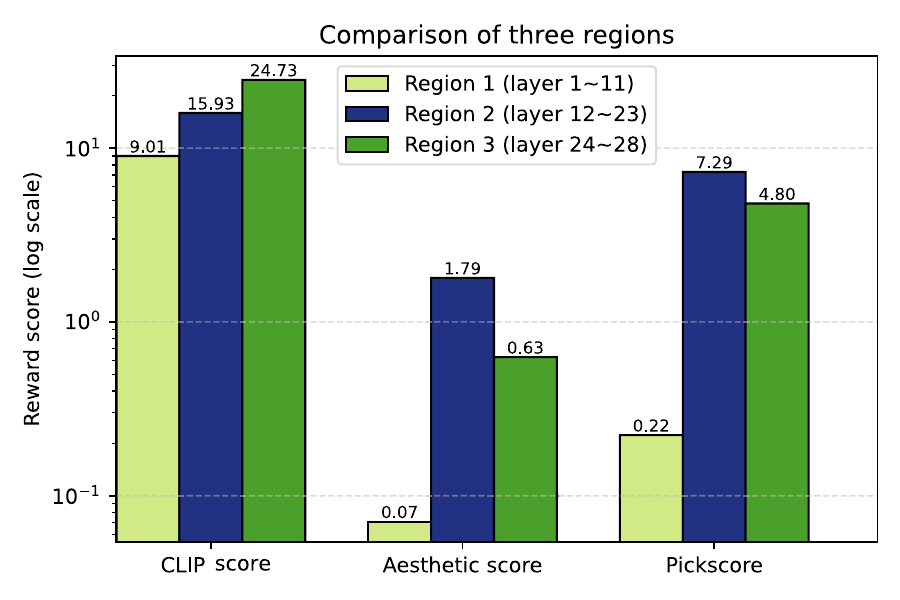}
   \caption{Comparison of the response degrees of reward scores to different regions. CLIP score is quite sensitive to changes in all three regions but is most significantly affected by the deep layer. Aesthetic score and Pickscore are most sensitive to changes in the middle layers but have little impact on the shallow layer. Therefore, we can influence the corresponding reward score by perturbing specific layers.}
   \label{fig:reward}
\end{figure}

With all layers used as conditions, we then dive into the relations and properties of individual VLM layers. We find adjacent layers exhibit higher similarity, and several regions with distinct intrinsic properties emerge, which aligns with different reward scores. \textbf{Relations}: We first extract the visual tokens from each layer after the above training and analyze their relationships using cosine similarity, as shown in \cref{fig:sim}. Similar adjacent phenomena have been observed in prior work \cite{li2025unifusion} but without analysis of the properties of each layer for reward scores. \textbf{Properties}: we therefore empirically remove layers corresponding to specific regions and measure the impact on three reward scores, as shown in \cref{fig:reward}. We find that reward scores align the hierarchical properties of these layer regions: middle-layer regions align aesthetics (Aesthetic score \cite{hentschel2022clip}) and human preferences (Pickscore \cite{kirstain2023pick}) metrics, while deep-layer regions align semantic (CLIP score \cite{radford2021learning}) metrics. Consequently, within a multi-layer conditioning framework, it is possible to design an optimization strategy to facilitate layer-wise reward alignment, thereby enhancing generative performance during the Reinforcement Learning (RL) stage.

\begin{figure*}[t]
  \centering
  \includegraphics[width=0.85\linewidth]{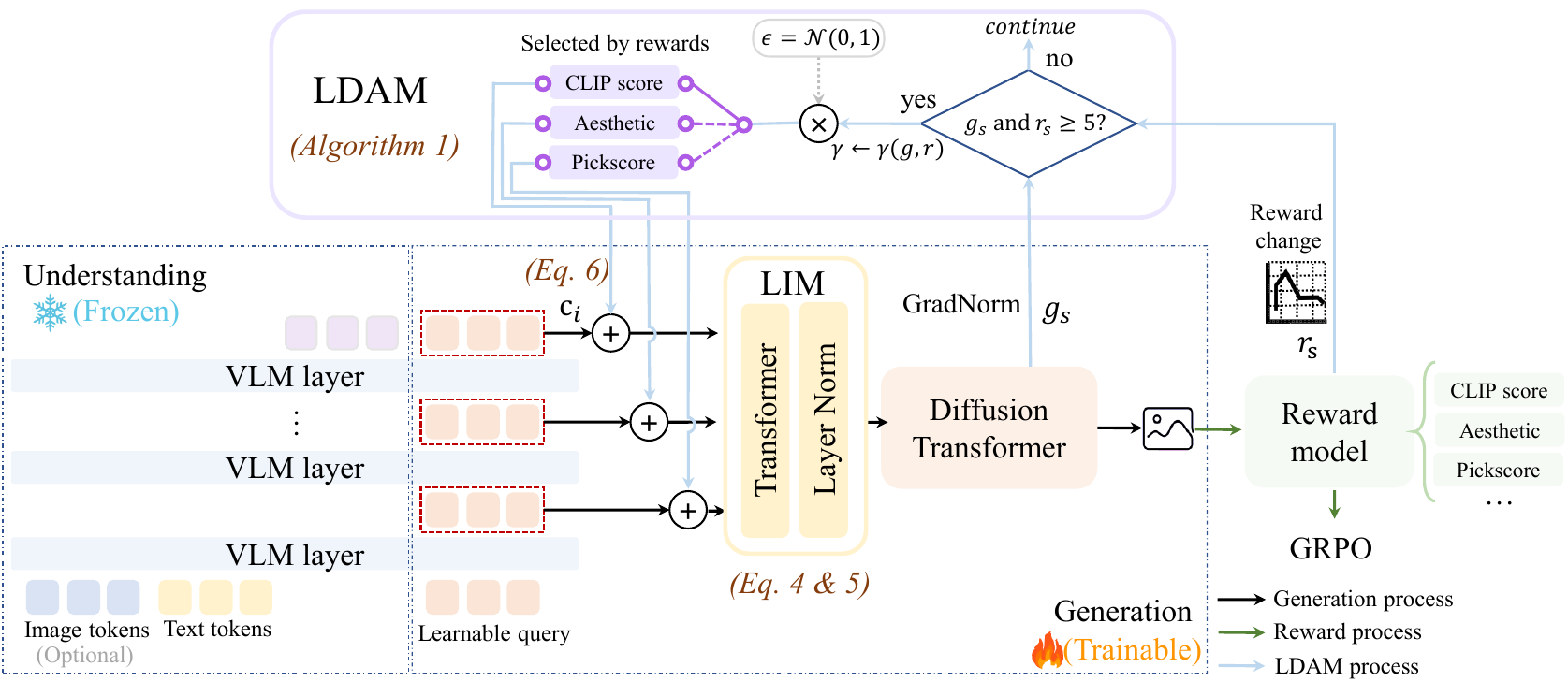}
   \caption{Overview of ParaUni. For understanding, after inputting image tokens and text tokens, the VLM performs understanding through an autoregressive process. For generation, the context information is compressed through a learnable query, and then the learnable queries from all layers are fed into a Layer Integration Module (LIM), which consists of a Transformer module and a Layer Norm layer. The integrated features are then sent into the cross attention in diffusion for generation. In the RL phase, we designed a Layer-wise Dynamic Adjustment Mechanism (LDAM), which adds Gaussian noise perturbation to specific layers through reward and GradNorm guidance. And different layers enable  perturbations under different rewards, as detailed in \cref{alg:cap}.}
   \label{fig:main}
\end{figure*}

In this paper, we propose a unified framework, dubbed \textbf{ParaUni}, that processes features from every layer of a VLM in parallel to bridge and enhance generation tasks with layer-wise guided RL. Concretely, as shown in \cref{fig:main}, our method employs a learnable-query–based unified architecture with learnable queries that extract visual tokens from each layer, which are then passed into the cross-attention in diffusion for the generation task. Our framework consists of two key components:
(1) \textbf{Layer Integration Module (LIM)}. We use a shared Transformer module to encode features from each layer in parallel, and then apply layer normalization to further align the different layer features. The processed features are finally fed into the diffusion model for generation.
(2) \textbf{Layer-wise Dynamic Adjustment Mechanism (LDAM)}. During the RL stage, when a specific reward is optimized, we perturb a specific layer's weights according to changes in the corresponding reward in \cref{fig:reward} and gradient norm to promote their improvement and stabilize training.
Extensive experiments show that ParaUni effectively improves generation, enhancing image detail and quality. Moreover, during the RL phase, it significantly boosts various reward metrics, demonstrating strong performance.

In summary, our contributions are as follows:

\begin{itemize}
\item We reveal that VLM layers encode properties from low-level detail to high-level semantics and that integrating visual tokens from these layers as conditions for diffusion substantially improves generation quality.
\item To make the best use of this characteristic, we design the LIM for parallel processing of VLM layers and introduce the LDAM with RL to further enhance performance.
\item Extensive experiments demonstrate that our method outperforms baselines and achieves state-of-the-art results. Our method enhances the details and quality of the generated images and at the same time promotes the improvement of multiple rewards in the RL stage.
\end{itemize}

\section{Related Works}
\label{sec:related}

\subsection{Unified Image Understanding and Generation}
Unified multimodal models~\cite{team2024chameleon, wu2024vila, xie2024show, wu2024janus, li2024synergen, deng2025bagel} aim to integrate visual understanding and generation within a single framework.  Previous work uses autoregressive VLMs~\cite{zhu2023minigpt, liu2023visualinstructiontuning, liu2024improved, li2024llava, yang2024qwen2, chen2024internvl, chen2024far, ma2025stage, zhang2025group, zhang2025maskfocus} for image understanding and diffusion models~\cite{2022LDM, 2023Pixelartalpha, 2023SDXL, 2024lumina, 2024pixartsigma, ramesh2021zero, ramesh2022hierarchical, dalle3} for image generation, creating an architectural gap that hinders unification. Some studies \cite{wang2024emu3, wu2024janus, ma2025janusflow, liu2025step1x} unify VLMs and diffusion models by feeding VLM's last hidden-layer features as conditioning to diffusion decoders. While leveraging both understanding and generation capabilities, simple concatenation limits information transfer. Other studies \cite{xie2024show, zhou2024transfusion, zhao2024monoformer} integrate diffusion's noise prediction into the autoregressive process, avoiding concatenation complexity and enabling information exchange. However, this causes mutual interference between tasks, reducing flexibility. Alternative paradigms \cite{deng2025bagel, shi2024llamafusion, li2025unifork} use separate parameters for each task to reduce conflicts but require substantial training resources. Some work using multi-layer VLM features \cite{li2025unifusion, cai2025hidream} also fails to leverage prior knowledge effectively, achieving suboptimal results.

\subsection{RL Algorithms for Image Generation}
Optimizing image generation models using reinforcement learning (RL) to align with human preferences has always been a key focus in generative tasks. \cite{xu2023imagereward} is the first work that optimizes diffusion models by using reward model scores. Diffusion-DPO \cite{wallace2024diffusion,yang2024using} introduces DPO into diffusion models, further improving the quality of generated images. With significant breakthroughs achieved by group relative policy optimization (GRPO) in large language models, RL for generation has gradually made rapid progress. \cite{liu2025flow} and \cite{xue2025dancegrpo} introduced randomness by reformulating the original ODE as an SDE, thereby incorporating GRPO into flow matching models. In addition, due to the instability of GRPO, some works \cite{wang2025pref, li2025mixgrpo} have proposed improvements, which further enhance the effectiveness of reinforcement learning.


\section{Method}
\label{sec:method}
\subsection{Preliminaries}
\textbf{Diffusion Models}. Diffusion models are a class of generative models that capture data distributions by inverting a forward corruption process, which is defined as: 
\begin{equation}
x_t = \alpha_t x_0 + \sigma_t \epsilon,\quad t \sim \mathcal{U}(0,1),\; \epsilon \sim \mathcal{N}(0,I),
\end{equation}
where $\alpha_t$ and $\sigma_t$ are monotonically decreasing and increasing functions of $t$, respectively. To reverse this process, a denoising network $\mathcal{F}_\theta$ is trained to predict the target $y$ at each time step $t$, conditioned on $c$ such as a class label or text prompt. In our setting, the condition $c$ is obtained from a VLM. This can be formulated as:
\begin{equation}
\mathcal{L}=\mathbb{E}_{x_0, c, \epsilon, t}[\Vert y-\mathcal{F}_\theta(x_t, c, t)\Vert_2^2],
\end{equation}
where the training target $y$ can be either the Gaussian noise $\epsilon$ in DDPM or the vector field $\epsilon-x_0$ in the flow model.

\textbf{RL for Diffusion Model}. 
Following the Flow-GRPO \cite{liu2025flow}, the flow model generates a set of $G$ candidate images $\{x_0^i\}_{i=1}^G$ along  with trajectories $\{x_T^i, x_{T-1}^i, \dots , x_0^i\}_{i=1}^G$ by reformulating trajectory sampling as a stochastic differential equation (SDE), injecting randomness into the process:
\begin{equation}
dx_t = \left( v_t + \frac{\sigma_t^2}{2t}(x_t + (1 - t)v_t) \right) dt + \sigma_t dw_t,
\end{equation}
which solves the problem of the original deterministic ODE formulation’s inability to generate diverse samples.
\begin{algorithm}[t]
\caption{Layer-wise Dynamic Adjustment Mechanism}\label{alg:cap}
\begin{algorithmic}[1]
\Require $Generation$ $model$ $\mathcal{G}$, $Reward$ $model$ $\mathcal{R}$.
\State \textbf{Initial:} $g_{n-1}, r_{n-1}, n_{cool}, r_s \gets inf$, 0, 0, 0
\For{$n \gets 1$ \textbf{to} $epochs$}
\State $g_n \gets Generation$ $model$ $\mathcal{G}$ 
\State $r_n \gets Reward$ $model$ $\mathcal{R}$ 
\If{$g_n \geq g_{n-1}*10^{2}$}:
    \State $g_{s} \gets True$ \Comment{Gradient norm guidance.}
\EndIf
\If{$r_n \leq r_{n-1}$}:
    \State $r_{s} \gets r_{s} + 1$ \Comment{Reward guidance.}
\EndIf
\If{$n_{cold} = 0$}:
    \If{$g_{s}$ and $r_{s} \geq 5$}:
        \State $\epsilon \sim \mathcal{N}(0,I)$
    
        \State $i \gets $Selected layer number \Comment{In \cref{sec:exp}}
        \State $\gamma \gets \gamma(g, r)$\Comment{In supplementary material}
        \State $c_i \gets c_i(1+\gamma\epsilon)$ \Comment{Core operation.}
        \State $r_s \gets 0$ \Comment{Restore to the original state.}
        \State $g_s \gets False$
        \State $n_{cool} \gets n$ \Comment{Cooling-off period begin.}
    \EndIf
\Else
    \State $n_{cool} \gets n_{cool}-1$ 
\EndIf
\State $g_{n-1} \gets g_{n}$ \Comment{Prepare for the next round.}
\State $r_{n-1} \gets r_{n}$
\EndFor
\end{algorithmic}
\end{algorithm}
\subsection{ParaUni}
In this section we introduce ParaUni, a unified architecture that simultaneously processes every layer of a VLM and feeds the integrated features into a diffusion model, improving generation quality for unified multimodal model.

\textbf{Overview}. Our framework follows the MetaQuery \cite{pan2025transfer} design and comprises $N$ learnable queries, a VLM, and a diffusion model. The VLM’s visual understanding capabilities remain unchanged, since its parameters are frozen. During image generation, the learnable queries extract textual information—or optionally visual information—during the VLM’s forward pass, and then pass that information to the diffusion model. In ParaUni, we feed the learnable queries from all layers as conditioning into a Layer Integration Module (LIM) to consolidate information across layers; the integrated representation is finally supplied to the diffusion model. During the RL stage, we propose a Layer-wise Dynamic Adjustment Mechanism (LDAM) that applies different optimization strategies to different layers according to different rewards, enabling improvement on multiple reward signals. In the sections below, we describe in detail how the LIM enhances generation quality and how the LDAM is designed for the RL stage.

\begin{figure}[t]
  \centering
  \includegraphics[width=1\linewidth]{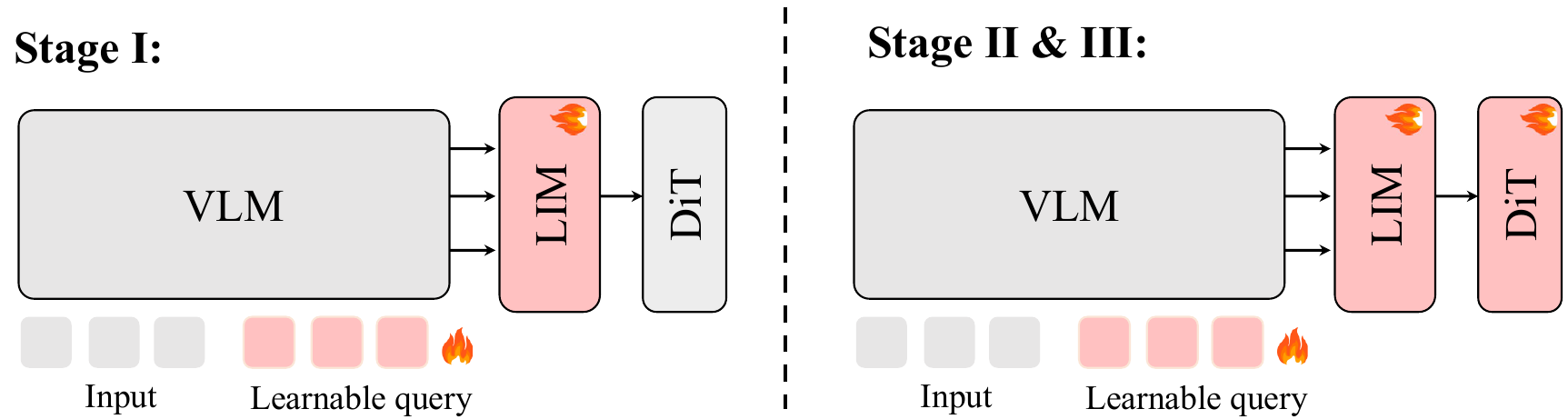}
   \caption{Illustration of the three training stages. Stage I is aligning VLM and diffusion, Stage II is fine-tuning with high-quality data, and Stage III uses RL to further enhance performance.}
   \label{fig:train}
\end{figure}

\begin{table*}[t]
    \centering
    
\caption{Quantitative results from the GenEval benchmark for text-to-image generation.}
    \resizebox{0.85\linewidth}{!}{
    \begin{tabular}{l l c c c c c c c}  
   \toprule
    \textbf{Type} & \textbf{Method} & \textbf{Single Obj.} & \textbf{Two Obj.} & \textbf{Counting} & \textbf{Colors} & \textbf{Position} & \textbf{Color Attri.} & \textbf{Overall$\uparrow$} \\
    \midrule
    \multirow{10}{*}{\textit{Gen. Only}} 
    & LlamaGen~\cite{sun2024autoregressive} & 0.71 & 0.34 & 0.21 & 0.58 & 0.07 & 0.04 & 0.32 \\
    & LDM~\cite{2022LDM} & 0.92 & 0.29 & 0.23 & 0.70 & 0.02 & 0.05 & 0.37 \\
    & SDv1.5~\cite{2022LDM} & 0.97 & 0.38 & 0.35 & 0.76 & 0.04 & 0.06 & 0.43 \\
    & PixArt-$\alpha$~\cite{chen2023pixart} & 0.98 & 0.50 & 0.44 & 0.80 & 0.08 & 0.07 & 0.48 \\
    & SDv2.1~\cite{2022LDM} & 0.98 & 0.51 & 0.44 & 0.85 & 0.07 & 0.17 & 0.50 \\
    & DALL-E 2~\cite{ramesh2022hierarchical} & 0.94 & 0.66 & 0.49 & 0.77 & 0.10 & 0.19 & 0.52 \\
    & Emu3-Gen ~\cite{wang2024emu3} & 0.98 & 0.71 & 0.34 & 0.81 & 0.17 & 0.21 & 0.54 \\
    & SDXL~\cite{2023SDXL}  & 0.98 & 0.74 & 0.39 & 0.85 & 0.15 & 0.23 & 0.55 \\
    & DALL-E 3~\cite{dalle3} & 0.96 & 0.87 & 0.47 & 0.83 & 0.43 & 0.45 & 0.67 \\
    & SD3-Medium~\cite{esser2024scaling} & 0.99&0.94& 0.72 & 0.89 & 0.33 & 0.60 & 0.74 \\
    \midrule
    \multirow{11}{*}{\textit{Unified}}
    & Chameleon~\cite{team2024chameleon} & - & - & - & - & - & - & 0.39 \\
    & SEED-X~\cite{ge2024seed} & 0.97 & 0.58 & 0.26 & 0.80 & 0.19 & 0.14 & 0.51 \\
    & LMFusion~\cite{shi2024llamafusion} & - & - & - & - & - & - & 0.63 \\
    & Show-o~\cite{xie2024show} & 0.95 & 0.52 & 0.49 & 0.82 & 0.11 & 0.28 & 0.68 \\
    & TokenFlow-XL~\cite{liu2024world} & 0.95 & 0.60 & 0.41 & 0.81 & 0.16 & 0.24 & 0.63 \\
    & Janus~\cite{wu2024janus} & 0.97 & 0.68 & 0.30 & 0.84 & 0.46 & 0.42 & 0.61 \\
    
    & MetaQuery~\cite{pan2025transfer} &  - & - & - & - & - & - & 0.74  \\
    
    & BLIP3-o-4B~\cite{chen2025blip3o} & - & - & - & - & - & - & 0.81 \\
    &BAGEL~\cite{deng2025bagel} & 0.99 &0.94 &\textbf{0.81} &0.88 &0.64 &0.63 &0.82\\
    &OpenUni\cite{wu2025openuni} & 0.99 & 0.92 & 0.76 & 0.91 & 0.82 & \textbf{0.77} & 0.86\\
    \cline{2-9}
    &ParaUni & \textbf{0.99} & \textbf{0.94} &0.78 &\textbf{0.91} &\textbf{0.83} &0.76 &\textbf{0.87}\\
    \bottomrule
  \end{tabular}
  }
  
  \label{tab:base}
\end{table*}


\textbf{Layer Integration Module}. Our early exploratory experiments reveal that the VLM’s layers encode rich, hierarchical representations. Therefore, instead of following the common practice of using only the final-layer features as conditioning, we introduce information from all layers into the diffusion model. Suppose we extract the learnable query $q_i$ from each layer $i$ of the VLM:
\begin{equation}
c_i = LN(f_\theta(q_i)), i\in[0, L],
\end{equation}
\begin{equation}
c = \frac{1}{n}(\sum_{i=1}^n(c_i)),
\end{equation}
where $L$ denotes the max layer number, $f_\theta$ denotes a Transformer module and $LN$ denotes a LayerNorm module. We use $c$ as the condition passed to DiT’s cross-attention, which produces the final high-quality image after denoising iterations.

\textbf{Layer-wise Dynamic Adjustment Mechanism}. Because different VLM layers respond differently to different rewards, we apply dynamic adjustments to specific layers during optimization of each reward. However, large adjustments can make training unstable. Inspired by prior work on inference-time scaling \cite{sadat2023cads}, which injects perturbations into text conditioning to improve generation quality and diversity, we perturb the layers that exhibit strong responses for a given reward when optimizing that reward. Concretely, we monitor each reward’s training signal in real time. If a reward shows a sustained decline, we use it as a guidance and inject noise sampled from a Gaussian distribution into the corresponding layers to encourage the model to explore a broader set of potential optima, expressed as follows:
\begin{equation}
c_i =c_i(1+\gamma\epsilon), \epsilon \sim \mathcal{N}(0,I),
\end{equation}
where $\gamma$ is a scale factor to control the degree of perturbations, defined in supplementary material.
Additionally, considering GradNorm as an indicator of training stability, we use it as a guidance for whether to apply the perturbation strategy. When GradNorm exhibits a large spike, we apply perturbations to the corresponding layer to encourage the model to escape the current unstable state. 
To ensure stable training, we have set a cooling-off period, which gradually increases with the number of training iterations.
During training, we first perform RL for one reward using the method described above. We then use the resulting checkpoint as the starting point for RL on the next reward, while preserving the corresponding layer perturbations. We repeat this process progressively to perform RL across multiple rewards. 
Finally, the detailed description of Layer-wise Dynamic Adjustment Mechanism is expressed as Algorithm 1.

\subsection{Training Recipe}
Our ParaUni training consists of three stages. First, we align the VLM and diffusion models, then fine-tune using high-quality data, and finally conduct RL, as shown in \cref{fig:train}.

\begin{table*}[t]
  \centering
  \caption{Quantitative Results from the DPG-Bench for Text-to-image generation. }
  \label{tab:dpg_bench}
  \resizebox{0.85\linewidth}{!}{
    \begin{tabular}{llccccccc}
      \toprule
      \textbf{Type} & \textbf{Method} & \textbf{Global} & \textbf{Entity} & \textbf{Attribute} & \textbf{Relation} & \textbf{Other} & \textbf{Overall$\uparrow$} \\
      \midrule
      \multirow{10}{*}{\textit{Gen. Only}} 
      & SDv1.5~\cite{2022LDM}  & 74.63 & 74.23 & 75.39 & 73.49 & 67.81 & 63.18 \\
      & PixArt-$\alpha$~\cite{chen2023pixart} & 74.97 & 79.32 & 78.60 & 82.57 & 76.96 & 71.11 \\
      & Lumina-Next~\cite{2024lumina}  & 82.82 & 88.65 & 86.44 & 80.53 & 81.82 & 74.63 \\
      & SDXL~\cite{2023SDXL}  & 83.27 & 82.43 & 80.91 & 86.76 & 80.41 & 74.65 \\
      & Playground v2.5~\cite{2024PG2.5}  & 83.06 & 82.59 & 81.20 & 84.08 & 83.50 & 75.47 \\
      & Hunyuan-DiT~\cite{2024hunyuandit}  & 84.59 & 80.59 & 88.01 & 74.36 & 86.41 & 78.87 \\
      & PixArt-$\Sigma$~\cite{2024pixartsigma}  & 86.89 & 82.89 & 88.94 & 86.59 & 87.68 & 80.54 \\
      & Emu3-Gen~\cite{wang2024emu3}  & 85.21 & 86.68 & 86.84 & 90.22 & 83.15 & 80.60 \\
      \midrule
      \multirow{7}{*}{\textit{Unified}}
      
      & Show-o~\cite{xie2024show}  & - & - & - & - & - & 67.27 \\
      & Janus~\cite{wu2024janus}  & 82.33 & 87.38 & 87.70 & 85.46 & 86.41 & 79.68 \\
      & Janus-Pro-1B~\cite{chen2025janus}  & 87.58 & 88.63 & 88.17 & 88.98 & 88.30 & 82.63 \\
      & MetaQuery~\cite{pan2025transfer}  & - & - & - & - & - & 80.04 \\
      & BLIP3-o-4B~\cite{chen2025blip3o}  & - & - & - & - & - & 79.36 \\
      &OpenUni~\cite{wu2025openuni} & 87.01 & \textbf{90.02} & \textbf{89.63} & 90.28 & \textbf{88.62} & 83.08\\
      \cline{2-8}
      & ParaUni  & \textbf{90.01} & 89.31 & 89.18 & \textbf{91.85} & 85.61 & \textbf{83.45} \\
      \bottomrule
    \end{tabular}
  }
\end{table*}

\textbf{Stage I}. Aligning the VLM and diffusion models. We follow the setup of OpenUni, in which the VLM is based on Intern-VL3-2B \cite{zhu2025internvl3exploringadvancedtraining} and the diffusion part is based on SANA-1.5-1.6B1024px \cite{xie2024sanaefficienthighresolutionimage}. The training data includes text-to-image-2M \cite{wang2025textatlas5m}, LAION-Aesthetic-6M \cite{schuhmann2022laion}, Megalith-10M \cite{meyer2024public}, RedCaps-5M \cite{desai2021redcaps}. 
In this stage, only the LIM module and learnable query are trained, while the other modules are frozen.

\textbf{Stage II}. We fine-tune the model using the high-quality data from BLIP3-o-60k \cite{chen2025blip3o}. In this stage, we train the parameters of the learnable query, the LIM module and the diffusion module. After this training stage, the model's performance has surpassed that of models trained only on the last layer with the same configuration.

\begin{figure*}[t]
  \centering
  \includegraphics[width=0.8\linewidth]{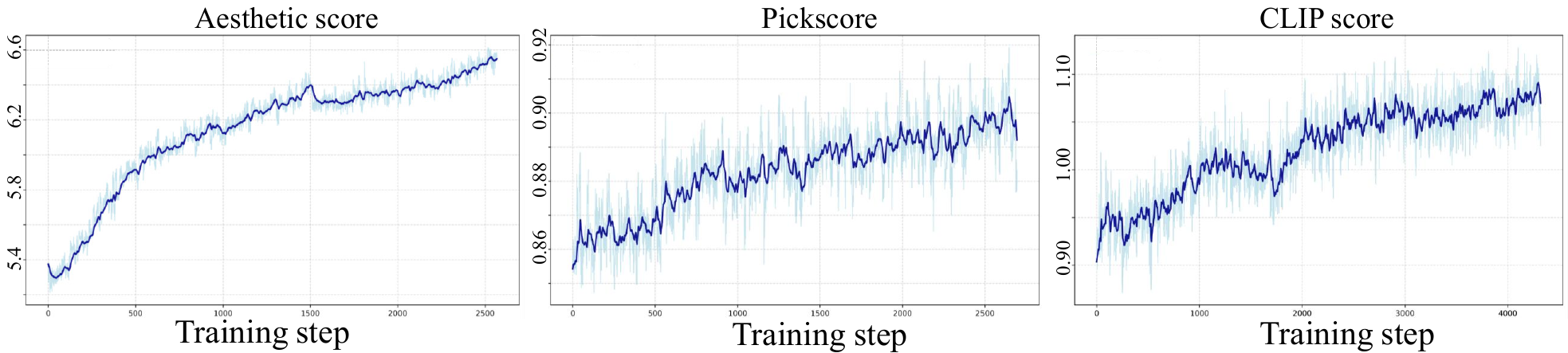}
   \caption{Reward change curves during the training process. As the training progresses, all reward scores gradually increase, demonstrating the effectiveness of our method.}
   \label{fig:socre}
\end{figure*}

\textbf{Stage III}. We conduct reinforcement learning training using multiple rewards following the Flow-GRPO \cite{liu2025flow} framework. The rewards include Aesthetic score, Pickscore and CLIP score. We first train using the Aesthetic score and Pickscore sequentially, then retain the weights of the corresponding layers, and finally train using the CLIP score. In this stage, the trainable parameters of the model are the same as in Stage II. More configuration will be detailed in the supplementary materials.

\textbf{Novelty Distinctions.} Considering prior work that also conditions diffusion models on all VLM layers, we now detail how ParaUni differs from those approaches. \cite{li2024playground} and \cite{cai2025hidream} perform one-to-one interactions between each VLM layer and each diffusion layer. Although this design leverages information from all VLM layers, it requires tight coupling between the VLM and the diffusion model, which reduces flexibility. \cite{li2025unifusion} is similar to our method in that it integrates multi-layer VLM features, but it does not analyze the distinct properties of those layers nor exploit those properties to accelerate or guide the RL process. In contrast, ParaUni explicitly studies layer-wise characteristics and uses that analysis to drive a Layer-wise Dynamic Adjustment Mechanism , allowing targeted perturbations and optimization strategies for different layers during RL while maintaining modularity and flexibility between VLM and diffusion model.

\section{Experiments}
\label{sec:exp}
\subsection{Implementation Details}
We train all three stages using the flow matching, which implements iterative denoising by predicting the velocity field. All models are optimized with the AdamW optimizer at a learning rate of 1e-4. The batch size is 512 and weight decay was set to 0.05. For the diffusion model, we adopt a cosine schedule. The VLM employs 256 learnable queries and 28 transformer layers. All experiments are conducted on NVIDIA A800 GPUs. In \cref{alg:cap}, for CLIP score, $i \in [24, 28]$; for Pickscore and Aesthetic score, $i \in [12, 23]$, which align with \cref{fig:reward}.

\subsection{Evaluation Benchmarks}
Because we do not train the VLM’s parameters, ParaUni’s understanding ability derives from the initial model InternVL3. The results for InternVL3’s understanding performance are provided in the supplementary material. To evaluate ParaUni’s generative capabilities, we primarily focus on the model’s adherence to prompts and the degree of semantic alignment. Specifically, we employ \textbf{GenEval} \cite{ghosh2024geneval}, which assesses a model’s ability to follow complex prompts and concentrates on generating images with correct object attributes, counts, positions, and colors. Results are reported across categories including single-object, multi-object, counting, color, and spatial position. We also use \textbf{DPG-Bench} \cite{hu2024ella} to examine the models’ complex semantic alignment for text-to-image generation under verbose and densely specified prompts. Finally, we assess the effectiveness of the RL stage by statistically evaluating reward scores, which include CLIP score for semantic alignment, Aesthetic score for image quality, and Pickscore reflecting human preference. Additionally, to evaluate ParaUni on image-to-image tasks, we report reconstruction performance using PSNR and SSIM on low-level details, as well as CLIP-based image similarity on high-level semantics in supplementary material.

\subsection{Comparison with the Baseline}
For baselines, we select purely generative models, including both diffusion- and autoregressive-based methods such as LlamaGen, SDv1.5, and DALL·E 2, as well as recent state-of-the-art unified understanding-and-generation models, such as Chameleon, Show‑o, BLIP3‑o, and BAGEL.

\begin{table}[t]
  \centering
  \caption{Ablation from the GenEval for text-to-image generation. }
  \label{tab:abl}
  \resizebox{0.9\linewidth}{!}{
    \begin{tabular}{lcccc}
      \toprule
      \textbf{Type} &  \textbf{Single Obj.}  & \textbf{Colors} & \textbf{Position} & \textbf{Overall$\uparrow$} \\
      \midrule
      (1) & 0.98  & 0.88 & 0.75 & 0.82 \\
      (2) & 0.99  & 0.90 & 0.81 & 0.85  \\
      (3) & 0.99  & 0.90 & 0.82 & 0.84  \\
      (4) & \textbf{1.00}  & 0.90 & 0.81 & 0.86  \\
      (5) & 0.98  & 0.61 & 0.75 & 0.73  \\
      (6) & 0.98  & 0.90 & 0.82 & 0.86  \\
      (7) & 0.98  & 0.90 & 0.81 & 0.86  \\
      \midrule
      Ours & 0.99  & \textbf{0.91} & \textbf{0.83} & \textbf{0.87}  \\
      \bottomrule
    \end{tabular}
  }
\end{table}

\begin{figure}[t]
  \centering
  \includegraphics[width=1\linewidth]{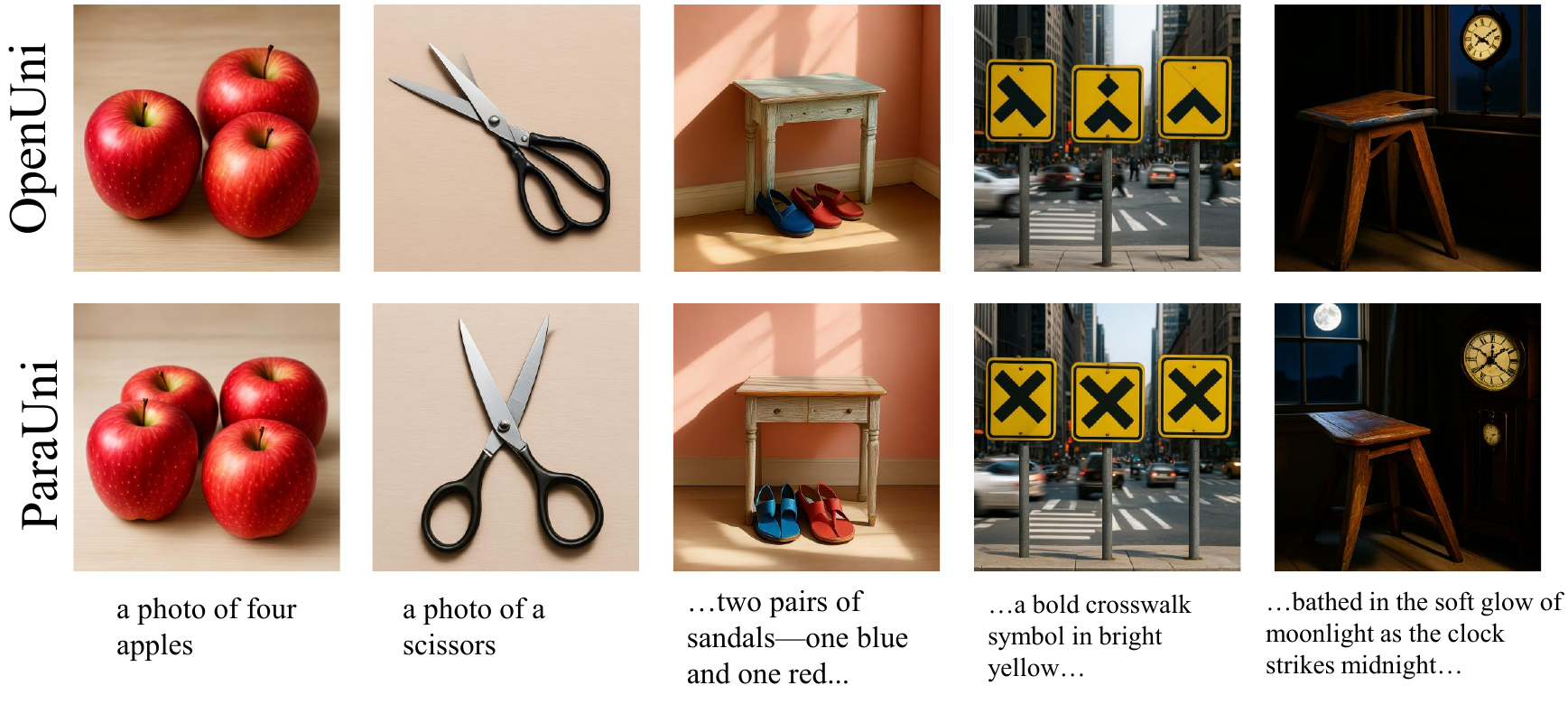}
   \caption{Qualitative comparison.}
   \label{fig:res}
\end{figure}

Table \ref{tab:base} compares our method against these baselines. As shown, our approach achieves a GenEval score of 0.87 and a DPG‑Bench score of 83.45, outperforming the unified understanding-and-generation baselines and substantially exceeding the purely generative models. This result indicates that unified understanding and generation architectures can markedly raise the performance ceiling of generative models relative to purely generative approaches. Furthermore, because we condition the generator on features from all VLM layers, we obtain an additional performance improvement over methods that use only a single layer as the conditioning signal.

\cref{fig:res} presents quantitative examples of outputs produced by our method. It illustrates that our approach effectively enhances both fine-grained visual detail and semantic alignment, yielding higher visual quality than models that condition on a single VLM layer, such as \cite{wu2025openuni}.

\subsection{Performance of RL}
\cref{fig:socre} is the effects of our method during the RL stage. The figure shows that each reward consistently increases over the course of training under our method, demonstrating robust generalization. Additionally, we provide quantitative results before and after RL and a comparison without our method in the supplementary materials.

\subsection{Ablation}
We conduct extensive ablation studies on the various modules of ParaUni. The results are shown in Table \ref{tab:abl}.

\textbf{Ablation on the Impact of Different Layer Selection Strategies on Generation}. We first evaluate the effect of conditioning on fewer visual-token layers, including (1) removing shallow, (2) removing middle, or (3) removing deep subsets, as well as (4) an interleaved scheme that uses every other layer. Table \ref{tab:abl} shows that omitting subsets of visual-token layers leads to degraded performance across metrics. Consequently, ParaUni conditions on \textbf{all} VLM layers.

\textbf{Ablations on the LIM}. We test alternative LIM configurations, including (5) removing LayerNorm module. Since removing the Transformer module would cause a significant drop in performance, it is not included in the ablation study. In each ablated variant in Table \ref{tab:abl}, performance declines relative to the full design, indicating that the components incorporated into our LIM each contribute positively to overall performance.

\textbf{Ablations on the LDAM}. We replace our proposed LDAM with alternative strategies, including (6) removing GradNorm-based guidance, (7) removing the reward degradation guidance. Removing either GradNorm guidance or the reward-degradation guidance led to worse training outcomes. Table \ref{tab:abl} shows that the LDAM design choices are important and effectively improve RL performance.

\textbf{Ablations on other baselines}. Our method is general and can serve as a plug-and-play module applied to other baselines. In Table \ref{tab:but2}, we also provide evaluation results on some relatively weaker baselines. As can be seen from Table \ref{tab:but2}, our method remains effective even when integrated into other baselines.

\begin{table}[t]
\centering
  \caption{Performance comparison of other baselines with ParaUni}
    \scalebox{0.9}{
\begin{tabular}{c|c c }
    \toprule
    Method & GenEval$\uparrow$  & DPG-Bench$\uparrow$
    \\
    \hline
    Janus-Pro (1B) & 0.73 & 82.63 \\
    Janus-Pro (1B)+ParaUni & \textbf{0.80} & \textbf{83.65} \\
    BLIP-3o (4B) & 0.81 & 79.36 \\
    BLIP-3o (4B)+ParaUni & \textbf{0.84} & \textbf{81.97} \\
    \bottomrule
  \end{tabular}
  }
\label{tab:but2}
\end{table}

\section{Conclusion}
\label{con}
In this paper, we propose ParaUni, a unified understanding-and-generation framework that conditions diffusion-based generation in parallel on information from all layers of a vision-language model. By leveraging multi-level representations across every VLM layer, we introduce a Layer Integration Module to aggregate these signals and thereby improve overall generation quality. We further devise a Layer-wise Dynamic Adjustment Mechanism to enhance multiple reward objectives during the reinforcement learning stage. Extensive experiments demonstrate that our method achieves state-of-the-art performance and exhibits strong practical potential.

\textbf{Acknowledgment.} This work was supported by the Anhui Provincial Natural Science Foundation under Grant 2108085UD12. We acknowledge the support of GPU cluster built by MCC Lab of Information Science and Technology Institution, USTC. The AI-driven experiments, simulations and model training were performed on the robotic AI-Scientist platform of Chinese Academy of Sciences.

{
    \small
    \bibliographystyle{ieeenat_fullname}
    \bibliography{main}
}


\end{document}